\documentclass[twoside]{article}

\usepackage[accepted]{aistats2023}
\usepackage[round]{natbib}

\usepackage[title,titletoc]{appendix}

\usepackage{array}
\usepackage{enumerate}
\usepackage{float} 
\usepackage{dsfont} 
\usepackage{mathrsfs} 
\usepackage{multirow}
\usepackage{hyperref}

\usepackage{amssymb}
\usepackage{amsmath}
\usepackage{bbm}

\usepackage{graphicx}

\usepackage{xcolor}
\usepackage{caption}

\usepackage{booktabs} 
\usepackage{multirow}

\usepackage{amsthm}
\newtheorem{thm}{Theorem}
\newtheorem{defn}{Definition}
\newtheorem{lem}{Lemma}

\newtheorem{asp}{Assumption}

\def\O{\mathcal{O}} 
\def\T{\mathcal{T}} 
 
\def\Z{\mathcal{Z}} 
\def\S{\mathcal{S}} 
\def\A{\mathcal{A}} 
\def\H{\mathcal{H}} 
\def\U{\mathcal{U}}

\usepackage{setspace} 
\usepackage{algorithm,algpseudocode} 

\usepackage{xcolor}         
\usepackage[colorinlistoftodos]{todonotes}

\newif\ifhighlight
\highlighttrue
\ifhighlight
\newcommand{\svbox}[1]{\todo[color=blue!40, inline]{\small (Sattar): #1}}
\newcommand{\mcbox}[1]{\todo[color=magenta!40, inline]{\small (Mark): #1}}
\newcommand{\abbox}[1]{\todo[color=red!40, inline]{\small (Alberto): #1}}

\newcommand{\sy}[1]{\textcolor{red}{#1}}
\else 
\newcommand{\svbox}[1]{}
\newcommand{\mcbox}[1]{}
\newcommand{\abbox}[1]{}
\newcommand{\sy}[1]{}

\newcommand{\sy}[1]{\textcolor{black}{#1}}
\fi 

\hypersetup{
    colorlinks=true,
    linkcolor=magenta,
    citecolor =blue,
    filecolor=magenta,      
    urlcolor=magenta,
}

\begin{document}
\runningauthor{
Sing-Yuan Yeh ~
Fu-Chieh Chang ~
Chang-Wei Yueh ~
Pei-Yuan Wu ~
Alberto Bernacchia ~
Sattar Vakili ~
}
%

%

\twocolumn[

\aistatstitle{Sample Complexity of Kernel-Based Q-Learning}

\aistatsauthor{
Sing-Yuan Yeh$^{1,2}$ ~
Fu-Chieh Chang$^{1,3}$ ~
Chang-Wei Yueh$^{4}$ ~
Pei-Yuan Wu$^{3,4}$ ~
Alberto Bernacchia$^{1}$ ~
Sattar Vakili$^{1}$ ~
}

\aistatsaddress{ 
\\
$^1$MediaTek Research \\ 
\vspace{-0.5em}
\\
$^2$Graduate Program of Data Science, National Taiwan University and Academia Sinica  \\
$^3$Graduate Institute of Communication Engineering, National Taiwan University \\
$^4$Department of Electrical Engineering, 
National Taiwan University 
} 
]

\begin{abstract}
Modern reinforcement learning (RL) often faces an enormous state-action space. Existing analytical results are typically for settings with a small number of state-actions, or simple models such as linearly modeled Q-functions. To derive statistically efficient RL policies handling large state-action spaces, with more general Q-functions, some recent works have considered nonlinear function approximation using kernel ridge regression. In this work, we derive sample complexities for kernel based Q-learning when a generative model exists. We propose a nonparametric Q-learning algorithm which finds an $\epsilon$-optimal policy in an arbitrarily large scale discounted MDP. The sample complexity of the proposed algorithm is order optimal with respect to $\epsilon$ and the complexity of the kernel (in terms of its information gain). To the best of our knowledge, this is the first result showing a finite sample complexity under such a general model.  

\end{abstract}


\section{Introduction}
In recent years, Reinforcement Learning (RL) has been successfully applied to several fields, including gaming \citep{silver2016mastering,lee2018deep,vinyals2019grandmaster}, autonomous driving \citep{kahn2017uncertainty}, microchip design \citep{mirhoseini2021graph}, robot control \citep{kalashnikov2018scalable}, and algorithm search \citep{fawzi2022discovering}. Real-world problems usually contain an enormous state-action space, possibly infinite. For example, the game of \emph{Go} has $10^{170}$ states \citep{silver2016mastering}, and the number of actions in the space of algorithms for matrix multiplication is $10^{30}$ \citep{fawzi2022discovering}. 
It is currently not fully understood how RL algorithms are able to learn successful policies to solve these problems.
Modern function approximators, such as kernel-based learning and deep neural networks, seem to be required for this success.

An important theoretical question is as follows. Consider a Markov decision process (MDP) with an unknown transition probability distribution. Suppose that a generative model~\citep{kakade2003sample} is available, which provides sample transitions from any state-action pair. How many samples are required to learn a sufficiently good policy? That is referred to as the sample complexity.

Previous works have derived theoretical bounds on the sample complexity, under certain simple settings such as tabular and linear MDPs. In the tabular setting, it was shown that the sample complexity of learning an $\epsilon$-optimal policy (that is the value function is at most $\epsilon$ away from the optimal value function) is in $\O(\frac{|\S||\A|}{\epsilon^2})$, where $|\S|$ and $|\A|$ are the cardinality of the state and action sets, respectively~\citep{kearns1998,gheshlaghi2013minimax, sidford2018near, sidford2018variance, agarwal2020model}, implying that for a very large state-action space, a virtually infinite number of samples is required to obtain a good performance.
Another line of work considers a linear MDP model, where the transition probability admits a linear representation in a $d$-dimensional state-action feature map.
It was shown that the sample complexity is in $\O(\frac{d}{\epsilon^2})$ in this case, that is independent of the size of state and action spaces~\citep{yang2019}.  Unfortunately, the linear
assumption is rather inflexible and not often the case in practice.

In order to address the limitations of small state-actions or simple models, arising from tabular and linear MDP assumptions, a few recent studies considered nonlinear function approximation over possibly infinite state-action domains using kernel ridge regression. In these works, the transition probability distribution (and sometimes the reward function) are flexibly represented using a kernel-based model~\citep{yang2020reinforcement,yang2020provably,yang2020}. The kernel-based models provide powerful regressor and uncertainty estimates, which can be leveraged to guide the RL algorithm. Furthermore, kernel-based models have a great representation capacity and can model a wide range of problems, considering that all continuous functions on compact subsets of $\mathbb{R}^d$ can be approximated using common kernels~\citep{srinivas2010}.
The existing works, however, do not address the specific question of sample complexity considered in this work, and instead derive regret bounds in the setting of an episodic MDP. A more detailed comparison is provided in Section~\ref{sec:related}. 

The kernel-based approaches may be understood as a linear model with an infinite-dimensional state-action feature map, that corresponds to, e.g., the Mercer eigenfeatures of the kernel (see Section~\ref{sec:RKHS}). In this sense, the linear model is a special case of the kernel-based model with a linear kernel. Nonetheless, the results on the sample complexity of linear MDPs do not extend to the kernel-based models, as those sample complexities scale with the dimension of the feature map (that is possibly infinite in the kernel setting). In contrast, in the kernel setting, the sample complexity depends on certain kernel-specific properties determined by the complexity of the kernel, which will be discussed in more detail. 
\subsection{Contributions}Considering a discounted MDP and the question of sample complexity~\citep[similar to][]{gheshlaghi2013minimax, sidford2018near,sidford2018variance, yang2019}, we extend and generalize the existing work as follows. 
\begin{itemize}
    \item We introduce Kernel-based Q-Learning, referred to as KQLearn, a sample collection algorithm, which returns an $\epsilon$-optimal policy with a finite sample complexity over a very general class of models.
    In comparison to tabular and linear MDP settings, KQLearn makes at least two innovative contributions. In the tabular setting, the samples are collected from all state-action pairs that leads to an $|\S||\A|$ scaling of the sample complexity. In the linear setting, the samples are collected from a set of state-actions spanning the entire state-action space (leading to the scaling of the sample complexity with dimension $d$). Then, an estimation of the parameters of the linear model are updated through value iteration. Neither of these approaches are feasible in our case with an infinite state-action space and a non-parametric kernel-based model. 
    KQLearn instead takes advantage of uncertainties provided by the kernel model to create a finite state-action set which is used for collecting the samples. These samples are then passed through an approximate Bellman operator using kernel ridge regression to update the value function (that is a continuous function over the entire state-action space).
    \item We derive a finite sample complexity for KQLearn under a wide range of kernel models. In particular, we consider two classes of kernels with exponentially ($\sigma_m\sim \exp(-m^{\beta_e})$, $\beta_e>0$) and polynomially ($\sigma_m\sim m^{-\beta_p}$, $\beta_p>1$) decaying Mercer eigenvalues $\sigma_m$ (see Definition~\ref{def:PolExp}). We prove a sample complexity of $\tilde{\O}(\frac{1}{\epsilon^2})$ and $\tilde{\O}\left((\frac{1}{\epsilon})^{\frac{2\beta_p}{\beta_p-1}}\right)$\footnote{The notations $\O$ and $\tilde{\O}$ are used to denote the mathematical order, and that up to hiding logarithmic factors, respectively.} under these two settings, respectively. To the best of our knowledge, this is the first finite sample complexity, for all $\beta_p>1$, and the first order optimal sample complexity in $\epsilon$,
    under the setting of polynomially decaying eigenvalues. Comparison with the related work is discussed in more detail in Section~\ref{sec:related}. As a special case, we recover the $\tilde{\O}(\frac{d}{\epsilon^2})$ sample complexity of the linear setting reported in~\cite{yang2019}.
\end{itemize}

We acknowledge that our bounds on the sample complexity of KQLearn may not be order optimal in the dependence on the discount factor $\gamma$. In particular, our bounds grow with $\frac{1}{(1-\gamma)^7}$ in the case of smooth kernels, similar to the PPQ-Learning algorithm proposed in~\cite{yang2019} for the linear setting. Under the tabular and linear settings, however, this dependency was improved to $\frac{1}{(1-\gamma)^3}$, in~\cite{sidford2018near} and~\cite{yang2019}, respectively. It appears a challenging problem whether the same improvement is feasible here. As mentioned above, even establishing a finite sample complexity is a challenging problem, and the sample complexities in the {existing} work may diverge with difficult kernels (some polynomial kernels as discussed in Section~\ref{sec:related}). 

\begin{table*}[t]
\begin{center}
\caption{The existing sample complexities under various settings, discussed in Section~\ref{sec:related}.}
\label{tab:related_works}
\begin{tabular}
{!{\vrule width 1.5pt}
>{\raggedright\arraybackslash}m{5.8cm}|
>{\centering \arraybackslash}m{1.5cm}|
>{\centering \arraybackslash}m{5.4cm}|
>{\centering \arraybackslash}m{3.1cm}
!{\vrule width 1.5pt}
}
\noalign{\hrule height 1.5pt}
Algorithm & 
MDP        & Setting       & Sample complexity 
\\ \hline
\noalign{\hrule height 1.5pt}
\citep[Q-learning with UCB,][]{jin2018q} &
Episodic &
 Tabular & 
 $\tilde{\mathcal{O}}\left(\frac{|\S| |\A| H^4 }{ \epsilon^2}\right)$  
 \\ \hline
 \citep[LSVI-UCB,][]{jin2019}  &
Episodic &
Linear
&
 $\tilde{\mathcal{O}}\left(\frac{d^3H^4}{\epsilon^2}\right)$  
\\ \hline
\multirow{2}{*}{\citep[KOVI,][]{yang2020}} &
\multirow{2}{*}{Episodic} &
Kernel-based, polynomial eigendecay &
$\tilde{\mathcal{O}}\left({H^4}(\frac{1}{\epsilon})^{\frac{2\beta_p}{\beta_p-2}}\right)$
\\ \cline{3-4}
 &
 &
Kernel-based, exponential eigendecay&
$\tilde{\mathcal{O}}\left(\frac{H^4}{\epsilon^2}\right)$
\\ \hline
 \citep[Variance-Reduced QVI]{sidford2018near} &
 Discounted  &
 Tabular
 & 
  $\tilde{\mathcal{O}}\left(\frac{|\S||\A|}{(1-\gamma)^3\epsilon^2}\right)$ 
 \\ \hline
 \citep[PPQ-Learning,][]{yang2019} &
 Discounted   &
 Linear
 &
 $\tilde{\mathcal{O}}\left(\frac{d}{(1-\gamma)^7\epsilon^2}\right)$   
 \\ \hline
 \citep[OPPQ-Learning,][]{yang2019} &
 Discounted   &
 Linear
 &
 $\tilde{\mathcal{O}}\left(\frac{d}{(1-\gamma)^3\epsilon^2}\right)$   
 \\ \hline
 \multirow{2}{*}{\textbf{KQLearn}} &
  \multirow{2}{*}{Discounted} & 
 Kernel-based, polynomial eigendecay
 &
$\tilde{\O}\left(
\frac{1}{\epsilon^{\frac{2\beta_p}{\beta_p-1}} (1-\gamma)^{\frac{7\beta_p-1}{\beta_p-1}}}
\right)~~~~~~$
\\ \cline{3-4}
 &
 & 
 Kernel-based, exponential eigendecay 
  & $\tilde{\O}\left(\frac{1}{\epsilon^{2}\left(1-\gamma\right)^7}\right)$
 \\

\noalign{\hrule height 1.5pt}
\end{tabular}
\end{center}
\end{table*}

\subsection{Related Work}\label{sec:related}

The specific problem of sample complexity in a discounted MDP using a generative model has been considered in tabular and linear settings. The results are summarized in Table~\ref{tab:related_works}. Other variants of the problem, consider MDPs in the absence of a generative model~\citep[e.g., see,][as representative works, as well as references therein]{azar2017minimax,jin2018q, jin2019,russo2019worst, yang2020provably, yang2020, kakade2020information, zhou2021provably, domingues2021kernel}, often episodic, with $T$ episodes of length $H$, and regret bounds depending on $T$ and $H$. The regret bounds can then be translated into sample complexities~\cite[e.g., see,][]{jin2018q, yang2020}. These results are also reported in Table~\ref{tab:related_works}.
Other approaches to nonlinear function approximation in RL include~models with
bounded \emph{eluder} dimension~\citep{wang2020provably, ayoub2020model} and smoothing kernels~\citep{domingues2021kernel}.
Among these works the two most relevant ones to ours are~\cite{yang2019} and~\cite{yang2020provably,yang2020}.

Similar to~\cite{yang2019}, we also consider sample complexity in a discounted MDP using a generative model. We consider a non-parametric kernel-based model, while they considered a parametric linear model. Thus, neither their algorithm nor their results extend to our setting. The linear setting is a special case of the kernel setting with a linear kernel, in which, we recover the $\tilde{\O}(\frac{d}{\epsilon^2})$ sample complexity, given in~\cite{yang2019}, for two algorithms: PPQ-Learning and OPPQ-Learning.  The latter improved the sample complexity with respect to the discount factor. 

Similar to~\cite{yang2020}, we also consider a kernel-based model. We consider sample complexity in a discounted MDP with a generative model, while they primarily considered regret bounds in an episodic MDP. They also reported sample complexities as a direct consequence of their regret bounds. Specifically, under the two settings of exponentially and polynomially decaying eigenvalues, their sample complexities translate to $\tilde{\O}(\frac{1}{\epsilon^2})$ and $\tilde{\O}\left((\frac{1}{\epsilon})^{\frac{2\beta_p}{\beta_p-2}}\right)$, respectively. Under the polynomial setting, their sample complexity bound is larger than ours. In addition, their sample complexity is not always finite and may diverge when $1<\beta_p\le2$, that includes many cases of interest. This suboptimality and possibly trivial result is a consequence of the superlinear (thus, trivial) regret bounds when $1<\beta_p\le2$. See~\cite{vakili2021open}, for a detailed discussion on the theoretical challenges related to this result.

For example, consider the Mat{\'e}rn family of kernels as one of the most commonly used~\citep{snoek2012practical, shahriari2015taking} and theoretically interesting~\citep{srinivas2010} family of kernels. For a Mat{\'e}rn kernel with smoothness parameter $\nu$ on a $d$ dimensional input domain, $\beta_p=1+\frac{2\nu}{d}$~\citep{yang2020}. That implies the sample complexity in~\cite{yang2020} diverges when $d>2\nu$ (that is often the case when using the Mat{\'e}rn kernel). We, however, emphasize that the discounted MDP with a generative model and the episodic MDPs are different settings, and cannot be compared directly. Nonetheless, we present the first always finite sample complexity under a very general setting covering all kernels with polynomially decaying eigenvalues.

Another related problem is the kernel-based bandit problem~\citep{srinivas2010}, which corresponds to a degenerate MDP with $|\S|=1$. The kernel-based bandit problem is a well studied problem with order optimal regret bounds~\citep{salgia2021domain, li2022gaussian} and sample complexities~\citep{vakili2021}. The lower bounds on sample complexities for the squared exponential (SE) and Mat{\'e}rn kernels are reported in~\citep{scarlett2017lower}, which have the same scaling with $\epsilon$ (up to logarithmic factors) as in our results, showing the order optimality of our sample complexities with $\epsilon$ (see Section~\ref{sec:Res}). 

\paragraph{Paper structure:} In Section~\ref{sec:prelim}, the problem is formalized, after an overview of the background on MDPs and kernel ridge regression. In Section~\ref{sec:ppq}, KQLearn is presented. The results are discussed in Section~\ref{sec:Res}. A high level analysis is provided in Section~\ref{sec:anal}, while the details are deferred to the appendix.

\section{Preliminaries}
\label{sec:prelim}
In this section, we overview the background on MDPs and kernel ridge regression. We then formally state the problem of sample complexity for Q-learning under this setting. 
\subsection{Discounted Markov Decision Process}\label{sec:DMDP}
A discounted Markov Decision Process (MDP) can be described by the tuple $M=(\mathcal{S}, \mathcal{A}, P, r, \gamma)$, where $\S$ is the state space, $\A$ is the action space, $\gamma\in(0,1)$ is the discount factor, $r : \S \times \A \rightarrow [0,1]$ is the reward function and $P(\cdot|s,a)$ is the transition probability distribution\footnote{We intentionally do note use the standard term \emph{transition kernel} for $P$, to avoid confusion with the term \emph{kernel} in kernel-based learning. } on $\S$ for the next state from state-action pair $(s,a)$. We use the notation $\Z=\S\times\A$ to denote the state-action space. Our results generally hold true for (possibly very large and) finite $\Z$ or certain infinite $\Z$. For correctness, we assume that $\Z$ is a compact {subset} of $ \mathbb{R}^d$. 



The goal is to find a (possibly random) policy $\pi: \mathcal{S} \rightarrow \mathcal{A}$, that maximizes the long-term expected reward, i.e., the value function,
$$
V^\pi(s):=\mathbb{E}\left[\sum_{t=0}^{\infty} \gamma^t r\left(s_t, \pi\left(s_t\right)\right) \mid s_0=s\right],
$$
where $s_t\sim P(\cdot|s_{t-1},\pi(s_{t-1}))$ forms the trajectory of the states.
It can be shown that~\citep[e.g., see][]{puterman2014markov}, under mild assumptions (e.g., continuity of $P$, compactness of $\Z$, and boundedness of $r$)  there exists an optimal policy $\pi^*$ which attains the maximal possible value $V^*$ at every state,
$$
\forall s \in \mathcal{S}: \quad V^*(s):=V^{\pi^*}(s) =\max _\pi V^\pi(s).
$$
To simplify the notation, for a value function $V:\S\rightarrow \mathbb{R}$, let
\begin{equation*}
[PV](s,a):=\mathbb{E}_{s'\sim P(\cdot|s,a)}[V(s')].
\end{equation*}
The Q-function, also sometimes referred to as the state-action value function, of a policy $\pi$, and the optimal Q-function are defined as
\begin{align*}
Q^\pi(s, a)&=r(s, a)+\gamma [PV^{\pi}](s,a) \text{, and} \\ Q^*(s,a)&=Q^{\pi^*}(s,a),
\end{align*}
respectively. 
The Bellman operator $\mathcal{T}: \mathbb{R}^{\mathcal{S}} \rightarrow \mathbb{R}^{\mathcal{S}}$ is defined as
$$
\forall s \in \mathcal{S}: \quad[\mathcal{T} V](s)=\max _{a \in \mathcal{A}}\left\{r(s, a)+\gamma [P V](s,a)\right\} .
$$


\paragraph{Sample complexity of $\epsilon$-optimal policies:}
An $\epsilon$-optimal policy is defined as follows. 
\begin{defn}
($\epsilon$-optimal policy) A policy $\pi$ is called $\epsilon$-optimal if it achieves near optimal values from any initial state as follows: 
\begin{equation*}
V^\pi(s) \geq V^*(s)-\epsilon, \quad \forall s \in \mathcal{S},
\end{equation*}
or equivalently $\left\|V^\pi-V^*\right\|_{\infty} \leq \epsilon$.
\end{defn}

We aim to learn $\epsilon$-optimal policies using a small number of samples. 
In this work, following~\cite{kearns1998,gheshlaghi2013minimax,sidford2018near, sidford2018variance, yang2019}, we
suppose that a generative model~\citep{kakade2003sample} is given where the RL algorithm is able to query transition samples $s'\sim P(\cdot|s,a)$
for any state-action pair $(s,a)\in\Z$. 
The \emph{sample complexity} of an RL algorithm is defined as the number of such samples used by the algorithm to obtain an $\epsilon$-optimal policy.

\subsection{RKHS and Kernel Ridge Regression}\label{sec:RKHS}

The existing work achieving finite sample complexity in the RL setting typically assumes a small state-action space or linearly modeled MDPs. These results can be generalized and extended using kernel-based learning. In particular, a natural approach is to use elements of a known reproducing kernel Hilbert space (RKHS) to model the transitions.  In this section, we overview RKHSs and kernel ridge regression. 

Let $K: \Z \times \Z \rightarrow \mathbb{R}$ be a known positive definite kernel with respect to a finite Borel measure. 
Let $\mathcal{H}_K$ be the RKHS induced by $K$, where $\H_K$ contains a family of functions defined on $\Z$. Let $\langle\cdot, \cdot\rangle_{\H_K}: \H_K \times \H_K \rightarrow \mathbb{R}$ and $\|\cdot\|_{\H_K}: \H_K \rightarrow \mathbb{R}$ denote the inner product and the norm of $\H_K$, respectively.
The reproducing property implies that for all $f\in\H_K$, and $z\in\Z$, $\langle f, K(\cdot,z)\rangle_{\mathcal{H}_K}=f(z)$. 
Without loss of generality, we assume $K(z,z)\leq 1$ for all~$z$. Mercer theorem implies, under certain mild conditions, $K$ can be represented using an infinite dimensional feature map: 
\begin{eqnarray}\label{eq:Mercer}
K(z,z')=\sum_{m=1}^{\infty}\sigma_m\psi_m(z)\psi_m(z'). 
\end{eqnarray}
A formal statement and the details are provided in Appendix~\ref{appx:rkhs}.

\paragraph{Kernel ridge regression:}
Kernel-based models provide powerful regressor and uncertainty estimators (roughly speaking, surrogate posterior variances) which can be leveraged to guide the RL algorithm. In particular, consider an unknown function $f\in\H_K$. Consider a set $\U_J=\{z_j\}_{j=1}^J\subset \Z$ of $J$ inputs. Assume $J$ noisy observations $\{Y(z_j)=f(z_j)+\epsilon_j\}_{j=1}^J$ are provided, where $\epsilon_j$ are i.i.d. zero mean sub-Gaussian noise terms. Kernel ridge regression provides the following regressor and uncertainty estimate, respectively~\citep[see, e.g.,][]{scholkopf2002learning},
\begin{align}
\hat{f}_{\U_J}(z) &= k^{\top}_{\U_J}(z)(K_{\U_J}+\lambda^2 I_J)^{-1}Y_{\U_J},\nonumber\\
\Sigma^2_{\U_J}(z)&=K(z,z)-k^{\top}_{\U_J}(z)(K_{\U_J}+\lambda^2 I_J)^{-1}k_{\U_J}(z), \label{eq:var}
\end{align}
where $k_{\U_J}(z)=[K(z,z_1),\dots, K(z,z_J)]^{\top}$ is a $J\times 1$ vector of the kernel values between $z$ and observations, $K_{\U_J}=[K(z_i,z_j)]_{i,j=1}^J$ is the $J\times J$ kernel matrix, $Y_{\U_J}=[Y(z_1), \dots, Y(z_J)]^{\top}$ is the $J\times 1$ observation vector, $I_J$ is the identity matrix of dimensions $J$, and $\lambda>0$ is a free regularization parameter.

\paragraph{Confidence intervals:}
The prediction and uncertainties provided by kernel ridge regression allow us to use standard confidence intervals in the algorithm and analysis. In particular, various results exist stating that with probability
at least $1-\delta$, the prediction function satisfies $|f(z)-\hat{f}_{\U_J}(z)|\le\beta(\delta)\Sigma_{\U_J}(z)$ (either for fixed $z$, or simultaneously for all $z$) where the confidence interval width multiplier $\beta(\delta)$ depends on the properties of the observation noise and the complexity of $f$ in terms of its RKHS norm~\citep{srinivas2010, abbasi2013online, vakili2021, vakili2022improved}. If the domain $\Z$ is finite, the uniform confidence bounds readily follow from a union
bound over the confidence intervals for a fixed $z$. For continuous domains,
a discretization argument is typically used considering the following continuity assumption. 
%
\begin{asp}\label{asp:disc}
For each $n\in\mathbb{N}$, there exists a discretization $\mathbb{Z}$ of $\Z$ such that, for any $f\in \H_K$ with $\|f\|_{\H_K}\le C_K$, we have $f(z) - f([z])\le \frac{1}{n}$, where $[z] = \arg\min_{ z'\in \mathbb{Z}}||z'-z||_{l^2}$ is the closest point in $\mathbb{Z}$ to $z$, and $|\mathbb{Z}|\le cC_K^dn^{d}$, where $c$ is a constant independent of $n$ and $C_K$.
\end{asp}
Assumption~\ref{asp:disc} is a technical and mild assumption that holds for typical kernels such as SE and Mat{\'e}rn~with $\nu>1$~\citep{srinivas2010, chowdhury2017kernelized, vakili2021}.

In our analysis, we use the following confidence interval for the RKHS elements. 

\begin{lem}[\cite{vakili2021, vakili2022improved}]\label{lem:confV}
Consider a fixed design of observation points where $\U_J$ is independent of the observation noise. When the noise terms are sub-Gaussian with parameter $R$~\footnote{A random variable $X$ is said to be sub-Gaussian with parameter $R$ if its moment generating function is bounded by that of a zero mean Gaussian with variance $R^2$. } and $\|f\|_{\H_K}\le C_K$, the following each hold uniformly in $z\in\Z$, with probability $1-\delta$,
\begin{eqnarray}\nonumber
f(z)&\le& \hat{f}_{\U_J}(z)+\beta(\delta)\Sigma_{\U_J}(z),\\
f(z)&\ge& \hat{f}_{\U_J}(z)-\beta(\delta)\Sigma_{\U_J}(z), 
\end{eqnarray}
where $\beta(\delta)=\O \left(C_K+\frac{R}{\lambda}\sqrt{d\log(\frac{JC_K}{\delta})}\right)$. 
\end{lem}

\paragraph{Maximal information gain:} 

It is useful for our analysis to define maximal information gain $\Gamma_{K,\lambda}$, that is a kernel specific complexity term. It allows us to bound the total uncertainty in the kernel model using results similar to elliptical potential lemma~\citep{carpentier2020elliptical}. In particular, let us define
\begin{eqnarray}
\Gamma_{K,\lambda}(J) = \sup_{\U\subset\Z, |\U|\le J}\frac{1}{2}\log\det\left(I_J+\frac{1}{\lambda^2}K_{\U}\right). 
\end{eqnarray}
Then, we have the following. 

\begin{lem}[\cite{srinivas2010}]\label{lem:uncertainty}
For any set $\U_J\subset\Z$, we have
\begin{eqnarray}
\sum_{j=1}^J\Sigma^2_{\U_{j-1}}(z_j)\le \frac{2}{\log(1+1/\lambda^2)}\Gamma_{K,\lambda}(J).
\end{eqnarray}
\end{lem}


\subsection{Problem Formulation}

Consider the discounted MDP described in Section~\ref{sec:DMDP}. 
We are interested in designing an algorithm with a small sample complexity which obtains an $\epsilon$-optimal RL policy, under the assumption that the transition probability distribution lives in the RKHS of a known kernel. Without loss of generality, we assume its RKHS norm is bounded by $1$. 
\begin{asp}
\label{asp:RKHSnormP}
Assume that the transition probability distribution satisfies,
\begin{equation*}
\forall s' \in \mathcal{S}: \quad\|P\left(s^{\prime} \mid \cdot, \cdot\right)\|_{\H_K} \le 1\,.
\end{equation*}
\end{asp}

This assumption is very flexible given the generality of the RKHSs. This is a standard assumption which is also used in~\cite{yang2020}. We do not make any explicit assumptions on the Q-function related to the policy.
Recall the definition of $PV$ from Section~\ref{sec:DMDP}.
In Lemma~\ref{lem:RKHS_PV}, we prove that for any $V:\S\rightarrow [0,\frac{1}{1-\gamma}]$, $\|PV\|_{\H_K}\le \frac{{c}}{1-\gamma}$, as a consequence of Assumption~\ref{asp:RKHSnormP}, that is essential for our analysis.

\paragraph{Some generic notation:}
For any real number $x$, and real numbers $a,b$,
the notation $\Pi_{[a, b]}[x]$ is used to denote the projection of $x$ onto $[a, b]$. For any integer $J$, $I_J$ denotes the $J\times J$ identity matrix, and $\mathbf{0}_J$ denotes the $J\times 1$ zero vector.




\section{Kernel Based Q-Learning}
\label{sec:ppq}

In this section, we present a novel kernel based Q-learning algorithm, referred to as KQLearn. 
Recall $Q(s,a)=r(s,a)+\gamma [PV](s,a)$. The transition probability distribution $P$ and the value function $V$ are both unknown to the algorithm. The algorithm, thus, recursively approximates $PV$, in rounds, using kernel ridge regression of $PV$ {from} the observations in the previous round. 
I.e., the algorithm performs updates based on an approximate Bellman operator using predictions for $PV$ provided by the kernel model. The samples are collected based on uncertainties for $PV$ in the kernel model. For this purpose, the algorithm first creates a maximum uncertainty set which is used to collect the samples. 

\paragraph{Maximum Uncertainty Set ($\U_J$):} The algorithm starts with creating a maximum uncertainty set with size $J$ referred to as $\U_J\subset \Z$. This set is created based on the uncertainties provided by the kernel model. In particular, each state-action is added to this set based on the following rule: choose the state-action with the highest uncertainty in the kernel model.
\begin{eqnarray}
(s_j,a_j) =\underset{(s,a)\in\Z}{\arg\max}~\Sigma^2_{{\U}_{j-1}}(s,a). 
\end{eqnarray}
Then, recursively, $\U_j=\U_{j-1}\cup\{(s_j,a_j)\}$, starting from $\U_0=\varnothing$. 
The set $\U_J$ is then used to collect samples from the generative model.

The algorithm proceeds in rounds indexed by $\ell=1,\dots, L$. Each round $\ell$ receives
noisy observations $Y^{(\ell-1)}_{\U_J}$ of $PV$ from the previous round, $\ell-1$. These observations are then used within kernel ridge regression to form a regressor of $PV$ over entire $\Z$, and obtain new observation $Y^{(\ell)}$. The observation vector is initialized to a zero vector $Y^{(0)}=\mathbf{0}_{J}$.


%
During each round $\ell$, for each state-action pair $(s_j,a_j)\in \U_J$, a transition state $s_j'\sim P(\cdot|s_j,a_j)$ is acquired from the generative model. The observation $Y^{(\ell)}(s_j,a_j)$ is then given as follows:
\begin{eqnarray}\nonumber
&&\hspace{-4em}Y^{(\ell)}(s_j,a_j)=\Pi_{[0,\frac{1}{1-\gamma}]}\max_{a\in\A}
\bigg\{r(s_j',a)\\
&&+\gamma k_{{\U}_J}(s_j',a)^\top (K_{\U_J}+\lambda^2I_J)^{-1}Y_{\U_J}^{(\ell-1)}\bigg\}.
\end{eqnarray}

The second term on the right hand side is the regressor in kernel ridge regression on $PV$, using $Y_{\U_J}^{(\ell-1)}$ as a vector of observations. In the analysis, we show a high probability bound on the error of this regression. The vector $Y_{\U_J}^{(\ell)}=[Y^{(\ell)}(s_1,a_1),\dots, Y^{(\ell)}(s_J,a_J)]^{\top}$ can be understood as updated noisy observations of $PV$ which is passed to the next round, $\ell+1$. 
By definition of the value function and the assumption of bounded rewards, it can be easily checked that $0\le V^*(s)\le \frac{1}{{1-\gamma}}$, for all $s\in\S$. We thus project the value of $PV$ on $[0,\frac{1}{1-\gamma}]$ interval.

KQLearn collects $N=JL$ samples in total.\footnote{For the simplicity of presentation, we assume $N=JL$. When $J$ does not divide $N$, we can ignore the samples in the last round.} A pseudo-code is provided in Algorithm~\ref{alg:kppq}.

After collecting all samples, the KQLearn algorithm returns an RL policy $\pi$ which selects the actions based on the following proxy $Q$-function:
\begin{equation}\label{eq:proxyQ}
\widehat{Q}^{{(L)}}(s,a)  =r(s,a) +\gamma k^\top_{{\U}_J}(s,a) (K_{\U_J}+\lambda^2I_J)^{-1}Y_{\U_J}^{(L)}.
\end{equation}

Specifically, when state $s$ is observed, the policy $\pi$ selects the action $\pi(s) = \arg\max_{a\in\A}\widehat{Q}^{{(L)}}(s,a)$. 
The second term on the right hand side is the kernel ridge regression of $PV$ using the observation in round $L$ of the KQlearn algorithm.

\begin{algorithm}
\caption{Kernel-based Q-learning (KQLearn)}
\label{alg:kppq}
\begin{flushleft}
\hspace*{\algorithmicindent} \textbf{Input}  Discounted MDP $M$, kernel $K$, regularization \linebreak\hspace*{\algorithmicindent} parameter $\lambda>0$, and $N>0$\\
\hspace*{\algorithmicindent} \textbf{Output} $\widehat{Q}^{(L)}:\Z\rightarrow \mathbb{R}$

\end{flushleft}
\begin{algorithmic}[1]
\State{Initialize $L,J\in\mathbb{N}$, $N=LJ$.}
\State{Initialize $Y^{(0)}=\mathbf{0}_{J}$ and the set $\U_0=\varnothing$.}
\ForAll{$j=1,\ldots,J$}
\State{Update the function $\Sigma_{\U_{j-1}}(\cdot)$ using Equation \ref{eq:var}.}
\State{Pick $(s_j,a_j)\leftarrow\arg \max_{(s,a)\in\Z}\;\Sigma^2_{\U_{j-1}}(s,a)$.}
\State{$\U_{j}\leftarrow\U_{j-1}\cup\{(s_j,a_j)\}$}.
\EndFor
\vspace{1em}
\ForAll{$\ell=1,\ldots ,L$} \Comment{round}
\ForAll{$j=1,\ldots,J$}
\State{Obtain a sample transition state
\linebreak\hspace*{\algorithmicindent} $~~~~~~~s'\sim P(\cdot|s_j,a_j)$}
\State{Update the $Y^{(\ell)}$ as follows.
\linebreak\hspace*{\algorithmicindent} 
~~~$Y^{(\ell)}(s_j,a_j)\leftarrow \Pi_{[0,\frac{1}{1-\gamma}]}\max_{a\in\A}\big\{r(s',a)$
\linebreak\hspace*{\algorithmicindent}\hspace*{\algorithmicindent}
~~~$+\gamma k_{\U_J}^\top(s',a)  \left(K_{\U_J}+\lambda I_J\right)^{-1} Y^{(\ell-1)}_{\U_J}  \big\}$}
\EndFor
\EndFor
\State $
\widehat{Q}^{{(L)}}(\cdot)  =r(\cdot)+\gamma k^\top_{{\U}_J}(\cdot) (K_{\U_J}+\lambda^2I_J)^{-1}Y_{\U_J}^{(L)}.$

\end{algorithmic}
\end{algorithm}

\section{Sample Complexity of KQLearn}\label{sec:Res}

In this section, we present our theoretical results. The following theorem establishes a bound on the error in the value function for the policy obtained in the KQLearn algorithm.

\begin{thm}
\label{thm:main}
Consider the discounted MDP described in Section~\ref{sec:DMDP}. Consider the KQLearn algorithm described in Section~\ref{sec:ppq}. Under Assumptions~\ref{asp:disc} and~\ref{asp:RKHSnormP}, with probability at least $1-\delta$, 
\begin{equation*}
\begin{aligned}
&\hspace{-0.8cm}\left\|V^{\pi}-V^\ast\right\|_{\infty}\leq \quad 2\beta(\delta) \left(\frac{\gamma}{1-\gamma}\right)^2\sqrt{\frac{2\Gamma_{K,\lambda}(J)}{J}}\\
&\hspace{1.8cm}+2\gamma^{L-1}\left(\frac{1}{1-\gamma}\right)^2,
\end{aligned}
\end{equation*}
where $\beta(\delta)=\O\left(\frac{c}{1-\gamma}+\frac{1}{2\lambda(1-\gamma)}\sqrt{d\log(\frac{J}{(1-\gamma)\delta})}\right)$ and $c$ is a constant given in Lemma~\ref{lem:RKHS_PV}.
\end{thm}

When $J$ and $L$ are sufficiently large, both terms in the upper bound become arbitrarily small.

\subsection{Sample Complexity}\label{sec:SC}

We can obtain explicit sample complexities for the KQLearn algorithm, using kernel specific bounds on $\Gamma_{K,\lambda}$, which depend on the decay rate of the Mercer eigenvalues of $K$. In particular, we define the following characteristic eigendecay profiles (which are similar to those outlined in~\cite{Bartlett2018, vakili2021information, yang2020}). 

\begin{defn}\label{def:PolExp}
[Polynomial and Exponential Eigendecay]\label{Def:PolExp}
Consider the Mercer eigenvalues $\{\sigma_m\}_{m=1}^\infty$ of $K$ as given in Equation\ref{eq:Mercer} in a decreasing order.
\begin{enumerate}[(i)]
    \item For some $C_p>0$, $\beta_p> 1$, $K$ is said to have a $(C_p,\beta_p)$ polynomial eigendecay, if for all $m\in \mathbb{N}$, we have $\sigma_m\le C_pm^{-\beta_p}$.
    \item For some $C_{e,1},C_{e,2},\beta_e>0$, $K$ is said to have a $(C_{e,1},C_{e,2},\beta_e)$ exponential eigendecay, if for all $m\in \mathbb{N}$, we have $\sigma_m\le C_{e,1}\exp(-C_{e,2}m^{\beta_e})$.
\end{enumerate}
\end{defn}


We are now ready to present explicit bounds on the sample complexity for the very general classes of kernels with polynomial and exponential decay of Mercer eigenvalues.

\begin{thm}
\label{thm:samplecomplexity}
Consider the discounted MDP described in Section~\ref{sec:DMDP}. Consider the KQLearn algorithm described in Section~\ref{sec:ppq}, with
$L=\Theta\left(\frac{\log\left(\epsilon(1-\gamma)^2\right)}{1-\gamma}\right)$ and $J=\frac{N}{L}$. Under Assumptions~\ref{asp:disc} and~\ref{asp:RKHSnormP}, KQLearn obtains an $\epsilon$-optimal policy with probability at least $1-\delta$, with a sample complexity at most

\begin{itemize}
    \item In the case of a kernel with $(C_p,\beta_p)$ polynomial eigendecay,
    \begin{eqnarray}
    N=\tilde{\O}\left(\frac{\left(\log(\frac{1}{\delta})\right)^{\frac{\beta_p}{\beta_p-1}}}{\epsilon^{\frac{2\beta_p}{\beta_p-1}}\left(1-\gamma\right)^{\frac{7\beta_p-1}{\beta_p-1}}}\right)
    \end{eqnarray}
    \item In the case of a kernel with $(C_{e,1},C_{e,2},\beta_e)$ exponential eigendecay,
    \begin{eqnarray}
    N=\tilde{\O}\left(\frac{\log(\frac{1}{\delta})}{\epsilon^{2}\left(1-\gamma\right)^7}\right)
    \end{eqnarray}
\end{itemize}
\end{thm}

A detailed expression including the implied constants and logarithmic factors in the $\tilde{\O}$ notation is provided in Appendix~\ref{appx:proofthm}.

\paragraph{Specific Kernels:}
Our bounds on the sample complexity can be specialized for various kernels where the eigendecay or bounds on $\Gamma_{K,\lambda}$ is known~\citep[such as the ones in][]{srinivas2010, vakili2021information, vakili2021}.
Specifically, for the Mat{\'e}rn and SE kernels, we have, respectively, $$N=\tilde{\O}\left(\frac{\left(\log(\frac{1}{\delta})\right)^{1+\frac{d}{2\nu}}}{\epsilon^{2+\frac{d}{\nu}}\left(1-\gamma\right)^{7+\frac{3d}{\nu}}}\right),~\text{and}~~N=\tilde{\O}\left(\frac{\log(\frac{1}{\delta})}{\epsilon^{2}\left(1-\gamma\right)^{7}}\right).$$

\subsection{{Optimality of the Sample Complexities} }

The sample complexities given above are order optimal with respect to $\epsilon$. We compare them to the lower bounds on the sample complexity for kernel bandits (that is a special case of our setting when $|\S|=1$). In particular, \cite{scarlett2017lower} proved $\Omega\left((\frac{1}{\epsilon})^{2+\frac{d}{\nu}}\right)$ and  $\Omega\left(\frac{1}{\epsilon^2}\right)$ sample complexities for the Mat{\'e}rn and SE kernels, respectively. Our results are the first finite sample complexities for the RL problem under a very general case which includes all kernels with polynomially decaying eigenvalues. 

In terms of the discount factor, our sample complexities scale with $\frac{1}{(1-\gamma)^7}$ in the case of smooth kernels~\citep[similar to the PPQ-Learning algorithm, in the linear setting,][]{yang2019}. In the tabular and linear settings, however, this has been improved to $\O(\frac{1}{(1-\gamma)^3})$. It remains an interesting problem for future investigation that whether the dependency of the sample complexity on the discount factor can be improved to $\O(\frac{1}{(1-\gamma)^3})$, also in the kernel setting. As discussed in the introduction, in the kernel setting, neither observing all state-actions nor a parametric update of the model through value iteration is feasible. Thus, a different approach to algorithm design and analysis is required that is increasingly more challenging among these settings: \emph{tabular$\rightarrow$ linear $\rightarrow$ kernel-based}.

\section{Analysis}\label{sec:anal}

Theorem~\ref{thm:samplecomplexity} is a consequence of Theorem~\ref{thm:main}, and using the kernel specific bounds on $\Gamma_{K,\lambda}$. The proof of Theorem~\ref{thm:main} builds on several components including tracking the approximation error in kernel ridge regression and convergence error of an approximate Bellman operator. In this section, we overview the main steps in the proof of Theorem~\ref{thm:main}, while deferring the details to the appendix. 

\paragraph{Approximate Bellman operator:} 

Recall the Bellman operator defined in Section~\ref{sec:DMDP}. The transition probability distribution and the value function are complex non-linear functions on continuous domains, unknown to the algorithm. We thus define an approximate Bellman operator $\widehat{\T}$, which uses noisy observations of $PV$ on a fixed set $\U_J$, and takes advantage of kernel ridge regression, to perform an approximate Bellman operator update. In particular, for all $V:\S\rightarrow [0, \frac{1}{1-\gamma}]$, and a fixed set $\U_J\subset \Z$, let us define
\begin{eqnarray}\nonumber
&&\hspace{-4em}[\widehat{\T}V](s) = \max_{a\in\A} \bigg\{r(s,a)\\
&&+\gamma k^{\top}_{\U_J}(s,a)(K_{\U_J}+\lambda^2 I_J)^{-1}[\widehat{P}V]_{\U_J}\bigg\},\label{eq:approxbell}
\end{eqnarray}
where $[\widehat{P}V](s,a)=V(s')$ is a random variable, $s'\sim P(\cdot|s,a)$ is a random transition state, and $$[\widehat{P}V]_{\U_J}=\bigg[[\widehat{P}V](s_1,a_1),\dots, [\widehat{P}V](s_J,a_J)]\bigg]^{\top}\;.$$  
In the KQLearn algorithm, define 
\begin{eqnarray}\nonumber
&&\hspace{-4em}\widehat{V}^{(\ell)}(s) = \max_{a\in\A}\bigg\{r(s,a)\\
&&+\gamma k_{\U_J}^\top(s,a)  \left(K_{\U_J}+\lambda^2 I_J\right)^{-1} Y^{(\ell)}_{\U_J}  \bigg\},
\end{eqnarray}
which we refer to as proxy value function (similar to the proxy Q-function given in~\eqref{eq:proxyQ}). 
We then have the following recursive relation over $\ell$.
\begin{eqnarray}
\widehat{V}^{(\ell)} = \widehat{\T}\Pi_{[0,\frac{1}{1-\gamma}]}[\widehat{V}^{(\ell-1)}].
\end{eqnarray}

\paragraph{Error in proxy value function:}
In order to bound the error in the value function of the policy $\pi$ obtained by KQLearn, $\|V^{\pi}-V^*\|$, we need to bound the error in the proxy Q-function given in~\eqref{eq:proxyQ}, which is used to obtain $\pi$. The error in proxy Q-function can be bounded based on the error in the proxy value function at round $L-1$. In particular, we have
\begin{eqnarray}
\|\widehat{Q}^{(L)}-Q^*\|_{\infty} \le \|\widehat{V}^{(L-1)}-V^*\|_{\infty}.
\end{eqnarray}

Therefore, we next bound the error in the proxy value function. We can write the error in proxy value function as the sum of two terms: the error in approximate Bellman operator and the error in the value function using true Bellman operator.
Specifically,
\begin{eqnarray}\nonumber
&&\hspace{-2em}\bigg\|\widehat{V}^{(L-1)} -V^*\bigg\|_{\infty}=
\bigg\|\widehat{\T}\Pi_{[0,\frac{1}{1-\gamma}]}\widehat{V}^{(L-2)}-V^*\bigg\|_{\infty}\\\nonumber
&&\le\bigg\|\widehat{\T}\Pi_{[0,\frac{1}{1-\gamma}]}\widehat{V}^{(L-2)} -\T\Pi_{[0,\frac{1}{1-\gamma}]}\widehat{V}^{(L-2)} \bigg\|_{\infty}\\
&&+ \bigg\|\T\Pi_{[0,\frac{1}{1-\gamma}]}\widehat{V}^{(L-2)} -V^*\bigg\|_{\infty}.
\end{eqnarray}

The second term can be recursively bounded which leads to the second term $\frac{2\gamma^{L-1}}{(1-\gamma)^2}$ in the error bound in Theorem~\ref{thm:main}. 
The first term leads to an important step in the analysis which is based on the error in kernel ridge regression. Specifically, let $V:\S\rightarrow [0,\frac{1}{1-\gamma}]$ be any value function. We have, for all $s\in\S$,
\begin{align}
&[\widehat{\T}V](s) -[\T V](s) \nonumber\\
&=\max_{a\in\A} \bigg\{r(s,a)+\gamma k^{\top}_{\U_J}(s,a)(K_{\U_J}+\lambda^2 I_J)^{-1}[\widehat{P}V]_{\U_J}\bigg\}\nonumber\\\nonumber
&\quad\quad- \max _{a \in \mathcal{A}}\left\{r(s, a)+\gamma [P V](s,a)\right\}\\\nonumber
&\le \gamma\max_{a\in\A}\bigg\{ k^{\top}_{\U_J}(s,a)(K_{\U_J}+\lambda^2 I_J)^{-1}[\widehat{P}V]_{\U_J}\\\label{eq:krrerrr}
&\quad\quad -[PV](s,a)\bigg\}.
\end{align}

\paragraph{Error in kernel ridge regression:}
The term inside $\max$ in Equation~\ref{eq:krrerrr} is the error in kernel ridge regression, 
where $PV$ is the target function, $[\widehat{P}V]_{\U_J}$ is a set of $J$ noisy observations, and $k^{\top}_{\U_J}(s,a)(K_{\U_J}+\lambda^2 I_J)^{-1}[\widehat{P}V]_{\U_J}$ is the regressor. In order to apply Lemma~\ref{lem:confV}, we need an upper bound on the RKHS norm of $PV$, as well as an upper bound on the sub-Gaussianity parameter of the observation noise in $\widehat{P}V$. These are established in the following lemmas. 
\begin{lem}\label{lem:RKHS_PV}
Consider an integrable value function $V:\S\rightarrow [0,\frac{1}{1-\gamma}]$.
Under Assumption~\ref{asp:RKHSnormP}, we have
\begin{eqnarray}
\|PV\|_{\H_K}\le \frac{c}{1-\gamma},
\end{eqnarray}
where $c=\int_{\S}V(s)ds$ is a constant determined by the {volume} of $\S$. 
\end{lem}
\begin{lem}\label{lem:sub-Gauss_PV}
Consider a transition probability distribution $P$, and an integrable value function $V:\S\rightarrow [0,\frac{1}{1-\gamma}]$. We have, for all $(s,a)$,
$\mathbb{E}\left[[\widehat{P}V](s,a)\right]=PV(s,a)$. In addition, $[\widehat{P}V](s,a)$ is a sub-Gaussian random variable with parameter $\frac{1}{2(1-\gamma)}$. 
\end{lem}

Lemma~\ref{lem:sub-Gauss_PV} follows from the definition of $\widehat{P}V$, as well as Hoeffding lemma for bounded random variables. A proof of Lemma~\ref{lem:RKHS_PV} is provided in Appendix~\ref{appx:prooflem}. 

Applying Lemma~\ref{lem:confV}, we obtain, with probability $1-\delta$, for all $(s,a)\in\Z$,
\begin{equation}\label{eq:kerbound}
\begin{aligned}
&\left|k^{\top}_{\U_J}(s,a)(K_{\U_J}+\lambda^2 I_J)^{-1}[\widehat{P}V]_{\U_J} - [PV](s,a)\right|\\
&\quad\quad\leq \beta(\delta)\Sigma_{\U_J}(s,a),
\end{aligned}
\end{equation}
where $\beta(\delta)=\O\left(\frac{c}{1-\gamma}+\frac{1}{2(1-\gamma)\lambda}\sqrt{d\log(\frac{Jc}{(1-\gamma)\delta})}\right)$.

Eventually, using Lemma~\ref{lem:uncertainty} on the total uncertainty, and by the design of $\U_J$, we bound $\Sigma_{\U_J}(s,a)$ on the right hand side. 

We thus bounded the two terms in~\eqref{eq:krrerr}, that bounds the error in proxy value function. 
More details on the proof of theorems and the proof of lemmas are provided in Appendix~\ref{appx:proofthm} and Appendix \ref{appx:prooflem}, respectively.

\section{Conclusion}

Modern RL often faces an enormous state-action space and complex models. We considered the question of sample complexity in a discounted MDP with a generative model under the kernel setting, furthering a line of research in the literature~\citep[e.g., see][]{kearns1998, azar2017minimax,sidford2018near, sidford2018variance, yang2019}. We introduced a novel kernel-based Q learning algorithm referred to as KQLearn and proved a
finite bound on its sample complexity for very general classes of kernels. 
That is to the best of our knowledge the first finite sample complexity result under the general kernel setting (including all kernels with polynomially decaying eigenvalues). In addition, compared to the lower bounds on the special case of the kernel bandit problem, our sample complexities are tight with respect to $\epsilon$ in finding an $\epsilon$-optimal policy. Our sample complexities, however, scale possibly suboptimally with respect to the discount factor, which remains an interesting open problem for future investigation.

\bibliography{AISTATS_bib}

\begin{thebibliography}{}

\bibitem[Abbasi-Yadkori, 2013]{abbasi2013online}
Abbasi-Yadkori, Y. (2013).
\newblock Online learning for linearly parametrized control problems.

\bibitem[Agarwal et~al., 2019]{agarwal2019}
Agarwal, A., Jiang, N., Kakade, S.~M., and Sun, W. (2019).
\newblock Reinforcement learning: Theory and algorithms.
\newblock {\em CS Dept., UW Seattle, Seattle, WA, USA, Tech. Rep}, pages 10--4.

\bibitem[Agarwal et~al., 2020]{agarwal2020model}
Agarwal, A., Kakade, S., and Yang, L.~F. (2020).
\newblock Model-based reinforcement learning with a generative model is minimax
  optimal.
\newblock In {\em Conference on Learning Theory}, pages 67--83. PMLR.

\bibitem[Ayoub et~al., 2020]{ayoub2020model}
Ayoub, A., Jia, Z., Szepesvari, C., Wang, M., and Yang, L. (2020).
\newblock Model-based reinforcement learning with value-targeted regression.
\newblock In {\em International Conference on Machine Learning}, pages
  463--474. PMLR.

\bibitem[Azar et~al., 2013]{gheshlaghi2013minimax}
Azar, M.~G., Munos, R., and Kappen, H.~J. (2013).
\newblock Minimax pac bounds on the sample complexity of reinforcement learning
  with a generative model.
\newblock {\em Machine learning}, 91(3):325--349.

\bibitem[Azar et~al., 2017]{azar2017minimax}
Azar, M.~G., Osband, I., and Munos, R. (2017).
\newblock Minimax regret bounds for reinforcement learning.
\newblock In {\em International Conference on Machine Learning}, pages
  263--272. PMLR.

\bibitem[Carpentier et~al., 2020]{carpentier2020elliptical}
Carpentier, A., Vernade, C., and Abbasi-Yadkori, Y. (2020).
\newblock The elliptical potential lemma revisited.
\newblock {\em arXiv preprint arXiv:2010.10182}.

\bibitem[Chatterji et~al., 2019]{Bartlett2018}
Chatterji, N., Pacchiano, A., and Bartlett, P. (2019).
\newblock Online learning with kernel losses.
\newblock In {\em Proceedings of Machine Learning Research}, volume~97, pages
  971--980, Long Beach, California, USA. PMLR.

\bibitem[Chowdhury and Gopalan, 2017]{chowdhury2017kernelized}
Chowdhury, S.~R. and Gopalan, A. (2017).
\newblock On kernelized multi-armed bandits.
\newblock In {\em International Conference on Machine Learning}, pages
  844--853. PMLR.

\bibitem[Christmann and Steinwart, 2008]{Christmann2008}
Christmann, A. and Steinwart, I. (2008).
\newblock {\em Support Vector Machines}.
\newblock Springer New York, NY.

\bibitem[Domingues et~al., 2021]{domingues2021kernel}
Domingues, O.~D., M{\'e}nard, P., Pirotta, M., Kaufmann, E., and Valko, M.
  (2021).
\newblock Kernel-based reinforcement learning: A finite-time analysis.
\newblock In {\em International Conference on Machine Learning}, pages
  2783--2792. PMLR.

\bibitem[Fawzi et~al., 2022]{fawzi2022discovering}
Fawzi, A., Balog, M., Huang, A., Hubert, T., Romera-Paredes, B., Barekatain,
  M., Novikov, A., R~Ruiz, F.~J., Schrittwieser, J., Swirszcz, G., et~al.
  (2022).
\newblock Discovering faster matrix multiplication algorithms with
  reinforcement learning.
\newblock {\em Nature}, 610(7930):47--53.

\bibitem[Jin et~al., 2018]{jin2018q}
Jin, C., Allen-Zhu, Z., Bubeck, S., and Jordan, M.~I. (2018).
\newblock Is q-learning provably efficient?
\newblock {\em Advances in Neural Information Processing Systems}, 31.

\bibitem[Jin et~al., 2020]{jin2019}
Jin, C., Yang, Z., Wang, Z., and Jordan, M.~I. (2020).
\newblock Provably efficient reinforcement learning with linear function
  approximation.
\newblock In {\em Conference on Learning Theory}, pages 2137--2143. PMLR.

\bibitem[Kahn et~al., 2017]{kahn2017uncertainty}
Kahn, G., Villaflor, A., Pong, V., Abbeel, P., and Levine, S. (2017).
\newblock Uncertainty-aware reinforcement learning for collision avoidance.
\newblock {\em arXiv preprint arXiv:1702.01182}.

\bibitem[Kakade et~al., 2020]{kakade2020information}
Kakade, S., Krishnamurthy, A., Lowrey, K., Ohnishi, M., and Sun, W. (2020).
\newblock Information theoretic regret bounds for online nonlinear control.
\newblock {\em Advances in Neural Information Processing Systems},
  33:15312--15325.

\bibitem[Kakade, 2003]{kakade2003sample}
Kakade, S.~M. (2003).
\newblock {\em On the sample complexity of reinforcement learning}.
\newblock University of London, University College London (United Kingdom).

\bibitem[Kalashnikov et~al., 2018]{kalashnikov2018scalable}
Kalashnikov, D., Irpan, A., Pastor, P., Ibarz, J., Herzog, A., Jang, E.,
  Quillen, D., Holly, E., Kalakrishnan, M., Vanhoucke, V., et~al. (2018).
\newblock Scalable deep reinforcement learning for vision-based robotic
  manipulation.
\newblock In {\em Conference on Robot Learning}, pages 651--673. PMLR.

\bibitem[Kearns and Singh, 1998]{kearns1998}
Kearns, M. and Singh, S. (1998).
\newblock Finite-sample convergence rates for q-learning and indirect
  algorithms.
\newblock In {\em Advances in Neural Information Processing Systems},
  volume~11. MIT Press.

\bibitem[Lee et~al., 2018]{lee2018deep}
Lee, K., Kim, S.-A., Choi, J., and Lee, S.-W. (2018).
\newblock Deep reinforcement learning in continuous action spaces: a case study
  in the game of simulated curling.
\newblock In {\em International Conference on Machine Learning,}, pages
  2937--2946. PMLR.

\bibitem[Li and Scarlett, 2022]{li2022gaussian}
Li, Z. and Scarlett, J. (2022).
\newblock Gaussian process bandit optimization with few batches.
\newblock In {\em International Conference on Artificial Intelligence and
  Statistics}, pages 92--107. PMLR.

\bibitem[Mercer, 1909]{Mercer1909}
Mercer, J. (1909).
\newblock Functions of positive and negative type, and their connection with
  the theory of integral equations.
\newblock {\em Philosophical Transactions of the Royal Society of London.
  Series A, Containing Papers of a Mathematical or Physical Character},
  209:415--446.

\bibitem[Mirhoseini et~al., 2021]{mirhoseini2021graph}
Mirhoseini, A., Goldie, A., Yazgan, M., Jiang, J.~W., Songhori, E., Wang, S.,
  Lee, Y.-J., Johnson, E., Pathak, O., Nazi, A., et~al. (2021).
\newblock A graph placement methodology for fast chip design.
\newblock {\em Nature}, 594(7862):207--212.

\bibitem[Puterman, 2014]{puterman2014markov}
Puterman, M.~L. (2014).
\newblock {\em Markov decision processes: discrete stochastic dynamic
  programming}.
\newblock John Wiley \& Sons.

\bibitem[Russo, 2019]{russo2019worst}
Russo, D. (2019).
\newblock Worst-case regret bounds for exploration via randomized value
  functions.
\newblock {\em Advances in Neural Information Processing Systems}, 32.

\bibitem[Salgia et~al., 2021]{salgia2021domain}
Salgia, S., Vakili, S., and Zhao, Q. (2021).
\newblock A domain-shrinking based bayesian optimization algorithm with
  order-optimal regret performance.
\newblock {\em Advances in Neural Information Processing Systems},
  34:28836--28847.

\bibitem[Scarlett et~al., 2017]{scarlett2017lower}
Scarlett, J., Bogunovic, I., and Cevher, V. (2017).
\newblock Lower bounds on regret for noisy gaussian process bandit
  optimization.
\newblock In {\em Conference on Learning Theory}, pages 1723--1742. PMLR.

\bibitem[Sch{\"o}lkopf et~al., 2002]{scholkopf2002learning}
Sch{\"o}lkopf, B., Smola, A.~J., Bach, F., et~al. (2002).
\newblock {\em Learning with kernels: support vector machines, regularization,
  optimization, and beyond}.
\newblock MIT press.

\bibitem[Shahriari et~al., 2015]{shahriari2015taking}
Shahriari, B., Swersky, K., Wang, Z., Adams, R.~P., and De~Freitas, N. (2015).
\newblock Taking the human out of the loop: A review of bayesian optimization.
\newblock {\em Proceedings of the IEEE}, 104(1):148--175.

\bibitem[Sidford et~al., 2018a]{sidford2018near}
Sidford, A., Wang, M., Wu, X., Yang, L., and Ye, Y. (2018a).
\newblock Near-optimal time and sample complexities for solving markov decision
  processes with a generative model.
\newblock {\em Advances in Neural Information Processing Systems}, 31.

\bibitem[Sidford et~al., 2018b]{sidford2018variance}
Sidford, A., Wang, M., Wu, X., and Ye, Y. (2018b).
\newblock Variance reduced value iteration and faster algorithms for solving
  markov decision processes.
\newblock In {\em Proceedings of the Twenty-Ninth Annual ACM-SIAM Symposium on
  Discrete Algorithms}, pages 770--787. SIAM.

\bibitem[Silver et~al., 2016]{silver2016mastering}
Silver, D., Huang, A., Maddison, C.~J., Guez, A., Sifre, L., Van Den~Driessche,
  G., Schrittwieser, J., Antonoglou, I., Panneershelvam, V., Lanctot, M.,
  et~al. (2016).
\newblock Mastering the game of go with deep neural networks and tree search.
\newblock {\em nature}, 529(7587):484--489.

\bibitem[Snoek et~al., 2012]{snoek2012practical}
Snoek, J., Larochelle, H., and Adams, R.~P. (2012).
\newblock Practical bayesian optimization of machine learning algorithms.
\newblock {\em Advances in neural information processing systems}, 25.

\bibitem[Srinivas et~al., 2010]{srinivas2010}
Srinivas, N., Krause, A., Kakade, S., and Seeger, M. (2010).
\newblock Gaussian process optimization in the bandit setting: No regret and
  experimental design.
\newblock In {\em ICML 2010 - Proceedings, 27th International Conference on
  Machine Learning}, pages 1015--1022.

\bibitem[Vakili et~al., 2021a]{vakili2021}
Vakili, S., Bouziani, N., Jalali, S., Bernacchia, A., and Shiu, D.-S. (2021a).
\newblock Optimal order simple regret for gaussian process bandits.
\newblock {\em Advances in Neural Information Processing Systems},
  34:21202--21215.

\bibitem[Vakili et~al., 2021b]{vakili2021uniform}
Vakili, S., Bromberg, M., Garcia, J., Shiu, D.-s., and Bernacchia, A. (2021b).
\newblock Uniform generalization bounds for overparameterized neural networks.
\newblock {\em arXiv preprint arXiv:2109.06099}.

\bibitem[Vakili et~al., 2021c]{vakili2021information}
Vakili, S., Khezeli, K., and Picheny, V. (2021c).
\newblock On information gain and regret bounds in gaussian process bandits.
\newblock In {\em International Conference on Artificial Intelligence and
  Statistics}, pages 82--90. PMLR.

\bibitem[Vakili et~al., 2021d]{vakili2021open}
Vakili, S., Scarlett, J., and Javidi, T. (2021d).
\newblock Open problem: Tight online confidence intervals for rkhs elements.
\newblock In {\em Conference on Learning Theory}, pages 4647--4652. PMLR.

\bibitem[Vakili et~al., 2022]{vakili2022improved}
Vakili, S., Scarlett, J., Shiu, D.-S., and Bernacchia, A. (2022).
\newblock Improved convergence rates for sparse approximation methods in
  kernel-based learning.
\newblock {\em arXiv preprint arXiv:2202.04005}.

\bibitem[Vinyals et~al., 2019]{vinyals2019grandmaster}
Vinyals, O., Babuschkin, I., Czarnecki, W.~M., Mathieu, M., Dudzik, A., Chung,
  J., Choi, D.~H., Powell, R., Ewalds, T., Georgiev, P., et~al. (2019).
\newblock Grandmaster level in starcraft ii using multi-agent reinforcement
  learning.
\newblock {\em Nature}, 575(7782):350--354.

\bibitem[Wang et~al., 2020]{wang2020provably}
Wang, R., Salakhutdinov, R., and Yang, L.~F. (2020).
\newblock Provably efficient reinforcement learning with general value function
  approximation.
\newblock {\em arXiv preprint arXiv:2005.10804}.

\bibitem[Yang and Wang, 2019]{yang2019}
Yang, L. and Wang, M. (2019).
\newblock Sample-optimal parametric q-learning using linearly additive
  features.
\newblock In {\em International Conference on Machine Learning}, pages
  6995--7004. PMLR.

\bibitem[Yang and Wang, 2020]{yang2020reinforcement}
Yang, L. and Wang, M. (2020).
\newblock Reinforcement learning in feature space: Matrix bandit, kernels, and
  regret bound.
\newblock In {\em International Conference on Machine Learning}, pages
  10746--10756. PMLR.

\bibitem[Yang et~al., 2020a]{yang2020provably}
Yang, Z., Jin, C., Wang, Z., Wang, M., and Jordan, M. (2020a).
\newblock Provably efficient reinforcement learning with kernel and neural
  function approximations.
\newblock {\em Advances in Neural Information Processing Systems},
  33:13903--13916.

\bibitem[Yang et~al., 2020b]{yang2020}
Yang, Z., Jin, C., Wang, Z., Wang, M., and Jordan, M.~I. (2020b).
\newblock On function approximation in reinforcement learning: Optimism in the
  face of large state spaces.
\newblock {\em arXiv preprint arXiv:2011.04622}.

\bibitem[Zhou et~al., 2021]{zhou2021provably}
Zhou, D., He, J., and Gu, Q. (2021).
\newblock Provably efficient reinforcement learning for discounted mdps with
  feature mapping.
\newblock In {\em International Conference on Machine Learning}, pages
  12793--12802. PMLR.

\end{thebibliography}
\addcontentsline{toc}{section}{References}


\newpage

\onecolumn

\newpage

\appendix

In the appendix, we provide some details and proofs omitted from the main paper due to space limit. In Appendix~\ref{appx:rkhs}, we provide a formal statement of Mercer theorem and a constructive definition of the RKHS. The proof of Theorems~\ref{thm:main} and~\ref{thm:samplecomplexity}, and auxiliary lemmas are provided in Appendix~\ref{appx:proofthm} and Appendix~\ref{appx:prooflem}, respectively. 

\section{Mercer theorem}
\label{appx:rkhs}
Mercer theorem \citep{Mercer1909} provides a representation of the kernel in terms of an infinite dimensional feature map (see, e.g. \cite{Christmann2008}, Theorem 4.49). Let $\mathcal{Z}$ be a compact metric space and $\mu$ be a finite Borel measure on $\mathcal{Z}$ (we consider Lebesgue measure in a Euclidean space). Let $L^2_\mu(\mathcal{Z})$ be the set of square-integrable functions on $\mathcal{Z}$ with respect to $\mu$. We further say a kernel is square-integrable if
\begin{equation*}
\int_{\mathcal{Z}} \int_{\mathcal{Z}} K(z, z')^2 \,d \mu(z) d \mu(z')<\infty.
\end{equation*}

\begin{thm}
(Mercer Theorem) Let $\mathcal{Z}$ be a compact metric space and $\mu$ be a finite Borel measure on $\mathcal{Z}$. Let $K$ be a continuous and square-integrable kernel, inducing an integral operator $T_K:L^2_\mu(\mathcal{Z})\rightarrow L^2_\mu(\mathcal{Z})$ defined by
\begin{equation*}
\left(T_K f\right)(\cdot)=\int_{\mathcal{Z}} K(\cdot, z') f(z') \,d \mu(z')\,,
\end{equation*}
where $f\in L^2_\mu(\mathcal{Z})$. Then, there exists a sequence of eigenvalue-eigenfunction pairs $\left\{(\sigma_m, \psi_m)\right\}_{m=1}^{\infty}$ such that $\sigma_m >0$, and $T_K \psi_m=\sigma_m \psi_m$, for $m \geq 1$. Moreover, the kernel function can be represented as
\begin{equation*}
K\left(z, z^{\prime}\right)=\sum_{m=1}^{\infty} \sigma_m \psi_m(z) \psi_m\left(z^{\prime}\right),
\end{equation*}
where the convergence of the series holds uniformly on $\mathcal{Z} \times \mathcal{Z}$.
\end{thm}

According to Mercer representation theorem (see, e.g., \cite{Christmann2008}, Theorem 4.51), the RKHS induced by $K$ can consequently be represented in terms of $\{(\sigma_m,\psi_m)\}_{m=1}^\infty$.

\begin{thm}(Mercer Representation Theorem) Let $\left\{\left(\sigma_m,\psi_m\right)\right\}_{i=1}^{\infty}$ be the Mercer eigenvalue eigenfunction pairs. Then, the RKHS of $K$ is given by
\begin{equation*}
\mathcal{H}_K=\left\{f(\cdot)=\sum_{i=1}^{\infty} w_i \sigma_i^{\frac{1}{2}} \psi_i(\cdot): w_i \in \mathbb{R},\|f\|_{\mathcal{H}_K}^2:=\sum_{i=1}^{\infty} w_i^2<\infty\right\}
\end{equation*}
\end{thm}
Mercer representation theorem indicates that the scaled eigenfunctions $\{\sqrt{\sigma_i}\psi_i\}_{i=1}^\infty$ form an orthonormal basis for $\H_K$.

\section{Proof of Theorems}\label{appx:proofthm}

In this section, we provide the proof of main theorems. 

\subsection{Proof of Theorem \ref{thm:main}.} 

The proof of Theorem~\ref{thm:main} builds on an approximate Bellman operator, that uses noisy observations of $PV$ within the rounds of KQLearn, and kernel ridge regression. We prove bounds on the error of this approximate Bellman operator. That is then used to bound the error in the value function of the policy obtained by KQLearn. 

\paragraph{Approximate Bellman operator:} 

Recall the Bellman operator defined in Section~\ref{sec:DMDP}. The transition probability distribution and the value function are complex non-linear functions on continuous domains, unknown to the algorithm. We thus define an approximate Bellman operator $\widehat{\T}$, which uses noisy observations of $PV$ on a fixed set $\U_J$, and takes advantage of kernel ridge regression, to perform an approximate Bellman operator update. In particular, for all $V:\S\rightarrow [0, \frac{1}{1-\gamma}]$, and a fixed set $\U_J\subset \Z$, let us define
\begin{eqnarray}\nonumber
&&\hspace{-4em}[\widehat{\T}V](s) = \max_{a\in\A} \bigg\{r(s,a)+\gamma k^{\top}_{\U_J}(s,a)(K_{\U_J}+\lambda^2 I_J)^{-1}[\widehat{P}V]_{\U_J}\bigg\},\label{eq:approxbell}
\end{eqnarray}
where $[\widehat{P}V](s,a)=V(s')$ is a random variable, $s'\sim P(\cdot|s,a)$ is a random transition state, and $$[\widehat{P}V]_{\U_J}=\bigg[[\widehat{P}V](s_1,a_1),\dots, [\widehat{P}V](s_J,a_J)]\bigg]^{\top}\;.$$  
In the KQLearn algorithm, define 
\begin{eqnarray}\nonumber
&&\hspace{-4em}\widehat{V}^{(\ell)}(s) = \max_{a\in\A}\bigg\{r(s,a)+\gamma k_{\U_J}^\top(s,a)  \left(K_{\U_J}+\lambda^2 I_J\right)^{-1} Y^{(\ell)}_{\U_J}  \big]\bigg\},
\end{eqnarray}
which we refer to as proxy value function (similar to the proxy Q-function given in~\eqref{eq:proxyQ}). 
We then have the following recursive relation over $\ell$.
\begin{eqnarray}
\widehat{V}^{(\ell)} = \widehat{\T}\Pi_{[0,\frac{1}{1-\gamma}]}[\widehat{V}^{(\ell-1)}].
\end{eqnarray}


\paragraph{Error in proxy value function:}
We next bound the error in the proxy value function. We can write the error in proxy value function as the sum of two terms: the error in approximate Bellman operator and the error in the value function using true Bellman operator.
Specifically, for $l>1$,
\begin{eqnarray}\nonumber
&&\hspace{-2em}\bigg\|\widehat{V}^{(l-1)} -V^*\bigg\|_{\infty}=
\bigg\|\widehat{\T}\Pi_{[0,\frac{1}{1-\gamma}]}\widehat{V}^{(l-2)}-V^*\bigg\|_{\infty}\\\nonumber
&&\le\underbrace{\bigg\|\widehat{\T}\Pi_{[0,\frac{1}{1-\gamma}]}\widehat{V}^{(l-2)} -\T\Pi_{[0,\frac{1}{1-\gamma}]}\widehat{V}^{(l-2)} \bigg\|_{\infty}}_{\text{Term I}}\\\label{eqap:twoterm}
&&~~~~~+ \underbrace{\bigg\|\T\Pi_{[0,\frac{1}{1-\gamma}]}\widehat{V}^{(l-2)} -V^*\bigg\|_{\infty}}_{\text{Term II}}.
\end{eqnarray}

We now bound the two terms on the right hand side of~\eqref{eqap:twoterm}.

\paragraph{Term I:} The first term leads us to an important step in the analysis which is based on the error in kernel ridge regression. Specifically, let $V:\S\rightarrow [0,\frac{1}{1-\gamma}]$ be any value function. We have, for all $s\in\S$,
\begin{align}\nonumber
[\widehat{\T}V](s) -[\T V](s) &=\max_{a\in\A} \bigg\{r(s,a)+\gamma k^{\top}_{\U_J}(s,a)(K_{\U_J}+\lambda^2 I_J)^{-1}[\widehat{P}V]_{\U_J}\bigg\}- \max _{a \in \mathcal{A}}\left\{r(s, a)+\gamma [P V](s,a)\right\}\\\label{eq:krrerr}
&\le \gamma\max_{a\in\A}\bigg\{ k^{\top}_{\U_J}(s,a)(K_{\U_J}+\lambda^2 I_J)^{-1}[\widehat{P}V]_{\U_J} -[PV](s,a)\bigg\}.
\end{align}

\paragraph{Error in kernel ridge regression:}
The term inside $\max$ in Equation~\ref{eq:krrerr} is the error in kernel ridge regression, 
where $PV$ is the target function, $[\widehat{P}V]_{\U_J}$ is a set of $J$ noisy observations, and $k^{\top}_{\U_J}(s,a)(K_{\U_J}+\lambda^2 I_J)^{-1}[\widehat{P}V]_{\U_J}$ is the regressor. In order to apply Lemma~\ref{lem:confV}, we need an upper bound on the RKHS norm of $PV$, as well as an upper bound on the sub-Gaussianity parameter of the observation noise in $\widehat{P}V$. These are established in Lemmas~\ref{lem:RKHS_PV} and~\ref{lem:sub-Gauss_PV}, respectively. Specifically, we have \begin{eqnarray}
\|PV\|_{\H_K}\le \frac{c}{1-\gamma}.
\end{eqnarray}
And, 
$[\widehat{P}V](s,a)$ is a sub-Gaussian random variable with parameter $\frac{1}{2(1-\gamma)}$.






Applying Lemma~\ref{lem:confV}, we obtain, with probability $1-\delta$, for all $(s,a)\in\Z$,
\begin{equation}\label{eq:kerbound}
\begin{aligned}
&\left|k^{\top}_{\U_J}(s,a)(K_{\U_J}+\lambda^2 I_J)^{-1}[\widehat{P}V]_{\U_J} - [PV](s,a)\right|\leq \beta(\delta)\Sigma_{\U_J}(s,a),
\end{aligned}
\end{equation}
where $\beta(\delta)=\O\left(\frac{c}{1-\gamma}+\frac{1}{2(1-\gamma)\lambda}\sqrt{d\log(\frac{Jc}{(1-\gamma)\delta})}\right)$.

\paragraph{Bounding $\Sigma_{\U_J}(s,a)$:}

Conditioning on a smaller subset of observation reduces the variance $\Sigma_{\U_j}(s,a)\ge \Sigma_{\U_J}(s,a)$, for all $j\leq J$ (due to positive definiteness of the kernel matrix). By the selection rule of the observation points:
\begin{eqnarray}
(s_j,a_j) =\underset{(s,a)\in\Z}{\arg\max}~\Sigma^2_{{\U}_{j-1}}(s,a),  
\end{eqnarray}
we have $\Sigma_{\U_{j-1}}(s,a)\le \Sigma_{\U_{j-1}}(s_j,a_j)$. 
Thus, for all $(s,a)\in\Z$, 
\begin{eqnarray}\nonumber
\Sigma^2_{\U_J}(s,a) &\le& \frac{1}{J} \sum_{j=1}^J \Sigma^2_{\U_{j-1}}(s,a)\\\nonumber
&\le& \frac{1}{J} \sum_{j=1}^J \Sigma^2_{\U_{j-1}}(s_j,a_j)\\\nonumber
&\le&\frac{2\Gamma_{K,\lambda}(J)}{\log(1+1/\lambda^2)J},
\end{eqnarray}
where the last line follows from Lemma~\ref{lem:uncertainty}. 

Replacing the bound on $\Sigma_{\U_J}(s,a)$, we obtain, for all $V:\S\rightarrow [0,\frac{1}{1-\gamma}]$, $s\in\S$, 
\begin{eqnarray}
[\widehat{\T}V](s) -[\T V](s) \le {\gamma}\beta(\delta)\sqrt{\frac{2\Gamma_{K,\lambda}(J)}{\log(1+1/\lambda^2)J}}. 
\end{eqnarray}

Thus, 
\begin{eqnarray}
\bigg\|\widehat{\T}\Pi_{[0,\frac{1}{1-\gamma}]}\widehat{V}^{(l-2)} -\T\Pi_{[0,\frac{1}{1-\gamma}]}\widehat{V}^{(l-2)} \bigg\|_{\infty}\le {\gamma}\beta(\delta)\sqrt{\frac{2\Gamma_{K,\lambda}(J)}{\log(1+1/\lambda^2)J}}.
\end{eqnarray}

\paragraph{Term II:}
We now bound the second term on the right hand side of~\eqref{eqap:twoterm}, by the contraction of Bellman operator.

\begin{eqnarray}\nonumber
\bigg\|\T\Pi_{[0,\frac{1}{1-\gamma}]}\widehat{V}^{(l-2)} -V^*\bigg\|_{\infty} &\le& \gamma\bigg\|\Pi_{[0,\frac{1}{1-\gamma}]}\widehat{V}^{(l-2)} - V^* \bigg\|_{\infty}\\
&\le& \gamma\bigg\|\widehat{V}^{(l-2)} - V^* \bigg\|_{\infty}. 
\end{eqnarray}
The second inequality follows from the observation that $V^*(s)\in[0,\frac{1}{1-\gamma}]$ for all $s\in\S$.

Combing the bounds on Term I and Term II, we obtain
\begin{eqnarray}\nonumber
&&\hspace{-2em}\bigg\|\widehat{V}^{(l-1)} -V^*\bigg\|_{\infty} \le  {\gamma}\beta(\delta)\sqrt{\frac{2\Gamma_{K,\lambda}(J)}{\log(1+1/\lambda^2)J}} + \gamma\bigg\|\widehat{V}^{(l-2)} - V^* \bigg\|_{\infty}. 
\end{eqnarray}

Recursively bounding the error in proxy value function at round $l$ using the error at round $l-1$ for $l=2,\dots, L-1$, we have, 

\begin{eqnarray}\nonumber
\bigg\|\widehat{V}^{(L-1)} -V^*\bigg\|_{\infty} &\le&  \beta(\delta)\sqrt{\frac{2\Gamma_{K,\lambda}(J)}{\log(1+1/\lambda^2)J}} \left(\sum_{i={1}}^{L-{1}} \gamma^{i}\right)+ \gamma^{L-1}\bigg\|\widehat{V}^{(0)} - V^* \bigg\|_{\infty}\\\nonumber
&\leq& \beta(\delta)\sqrt{\frac{2\Gamma_{K,\lambda}(J)}{\log(1+1/\lambda^2)J}} \left(\sum_{i=1}^{L-1} \gamma^{i}\right)+\frac{\gamma^{L-1}}{1-\gamma}, 
\end{eqnarray}
where the second inequality comes from $\bigg\|\widehat{V}^{(0)} - V^* \bigg\|_{\infty}\le \frac{1}{1-\gamma}$. 

Recall the definition of the proxy Q-function $\widehat{Q}^{{(L)}}$ given in~\eqref{eq:proxyQ}. We bound the error in $\widehat{Q}^{{(L)}}$ as follows. For all $(s,a)\in\Z$,
\begin{align*}
\left\|\widehat{Q}^{(L)}(s,a)-Q^*(s,a)\right\|_\infty &= \left\|\left(r(s,a) +\gamma k^\top_{{\U}_J}(s,a) (K_{\U_J}+\lambda^2I_J)^{-1}Y^{(L)}_{\U_J}\right) - \big(r(s,a)+\gamma[PV^\ast](s,a)\big)\right\|_\infty\\
&\leq\left\|\gamma \left(k^\top_{{\U}_J}(s,a) (K_{\U_J}+\lambda^2I_J)^{-1}[\widehat{P}\Pi_{[0,\frac{1}{1-\gamma}]}\widehat{V}^{(L-1)}]_{\U_J}-[P\Pi_{[0,\frac{1}{1-\gamma}]}\widehat{V}^{(L-1)}](s,a)\right)\right\|_\infty\\
&\quad\quad\quad+\left\|\gamma\left([P\Pi_{[0,\frac{1}{1-\gamma}]}\widehat{V}^{(L-1)}](s,a)-[PV^*](s,a)\right)\right\|_\infty\\
&\leq\gamma\beta(\delta)\sqrt{\frac{2\Gamma_{K,\lambda}(J)}{\log(1+1/\lambda^2)J}}+\gamma\left\|\mathbb{E}_{s'\sim P(\cdot|s,a)}[\Pi_{[0,\frac{1}{1-\gamma}]}\widehat{V}^{(L-1)}]-\mathbb{E}_{s'\sim P(\cdot|s,a)}[V^\ast]\right\|_\infty\\
&\leq\gamma\beta(\delta)\sqrt{\frac{2\Gamma_{K,\lambda}(J)}{\log(1+1/\lambda^2)J}}+\gamma\left\|\Pi_{[0,\frac{1}{1-\gamma}]}\widehat{V}^{(L-1)}-V^\ast\right\|_\infty\\
&\leq\gamma\beta(\delta)\sqrt{\frac{2\Gamma_{K,\lambda}(J)}{\log(1+1/\lambda^2)J}}+\gamma\left\|\widehat{V}^{(L-1)}-V^\ast\right\|_\infty\\
&\leq\gamma\beta(\delta)\sqrt{\frac{2\Gamma_{K,\lambda}(J)}{\log(1+1/\lambda^2)J}}+\gamma\left( \beta(\delta)\sqrt{\frac{2\Gamma_{K,\lambda}(J)}{\log(1+1/\lambda^2)J}}\left(\sum_{i=1}^{L-1} \gamma^{i}\right)+\frac{\gamma^{L-1}}{1-\gamma}\right)\\\nonumber
&= \beta(\delta)\sqrt{\frac{2\Gamma_{K,\lambda}(J)}{\log(1+1/\lambda^2)J}}\left(\sum_{i=1}^{L} \gamma^{i}\right)+\frac{\gamma^L}{1-\gamma}\\
&\le \frac{\gamma \beta(\delta)}{1-\gamma}\sqrt{\frac{2\Gamma_{K,\lambda}(J)}{\log(1+1/\lambda^2)J}}+\frac{\gamma^{L}}{1-\gamma}.
\end{align*}

The first inequality is a result of triangle inequality after adding and subtracting the term $\gamma[P\Pi_{[0,\frac{1}{1-\gamma}]}\widehat{V}^{(L-1)}](s,a)$. The second inequality is the error in kernel ridge regression bounded above in Term I. The third inequality bounds the difference in expectations with maximum difference. The fourth inequality is a consequence of the observation that $V^*\in[0,\frac{1}{1-\gamma}]$ interval. The fifth inequality is obtained using the bound on the error in the proxy value function given above.


\paragraph{The value function of $\pi$ obtained from the proxy Q-function:} The following lemma establishes that the error in the value function of $\pi$ can be bounded using the error in the proxy Q-function. 
\begin{lem}
\label{lem:valerr}
Consider any Q-function $\tilde{Q}:\S\times\A\rightarrow [0,\frac{1}{1-\gamma}+1]$ satisfying $\|\tilde{Q}-Q^\ast\|_\infty\leq\epsilon$.
Define policy $\pi_{\tilde{Q}}$ such that $\pi_{\tilde{Q}}(s)=\underset{a\in\A}{\arg\max}\;\tilde{Q}(s,a)$. 
Then, for all $s\in\S$,
\begin{equation*}
V^\ast(s)-V^{\pi_{\tilde{Q}}}(s)\leq \frac{2\epsilon}{1-\gamma}\,.
\end{equation*}
\end{lem}

Applying Lemma~\ref{lem:valerr} to $\widehat{Q}$ returned by KQLearn, it follows that with probability $1-\delta$, 
\begin{align*}
\left\|V^{\pi}-V^*\right\|_\infty \le \frac{2\gamma \beta(\delta)}{(1-\gamma)^2}\sqrt{\frac{2\Gamma_{K,\lambda}(J)}{\log(1+1/\lambda^2)J}}+\frac{2\gamma^L}{(1-\gamma)^2}.
\end{align*}
That completes the proof.

\subsection{Proof of Theorem \ref{thm:samplecomplexity}}

Theorem~\ref{thm:samplecomplexity} is a consequence of Theorem~\ref{thm:main} and the kernel specific bounds on $\Gamma_{K,\lambda}$. Several works have established bounds on~$\Gamma_{K,\lambda}$ for various kernels~\citep{srinivas2010, vakili2021information, vakili2021uniform}. We use the reuslt in \cite{vakili2021information} for kernels with polynomial and exponential eigendecay.

\paragraph{Polynomial Eigendecay:} Consider a kernel with $(C_p,\beta_p)$ polynomial eigendecay. We have the following bound on $\Gamma_{K,\lambda}$~\citep{vakili2021information}. For all $J\in\mathbb{N}$,
\begin{eqnarray}
\Gamma_{K,\lambda}(J) = \O(J^{\frac{1}{\beta_p}}\log^{1-\frac{1}{\beta_p}}(J))
\end{eqnarray}

We replace this bound in the error in the value function of $\pi$ obtained by KQLearn, in Theorem~\ref{thm:main}
\begin{eqnarray}\nonumber
\left\|V^{\pi}-V^\ast\right\|_{\infty}&\leq&  \frac{2\gamma\beta(\delta)}{(1-\gamma)^2}\sqrt{\frac{2\Gamma_{K,\lambda}(J)}{\log(1+\frac{1}{\lambda^2})J}} + \frac{2\gamma^L}{(1-\gamma)^2}.
\end{eqnarray}
We then obtain
\begin{eqnarray}\nonumber
\left\|V^{\pi}-V^\ast\right\|_{\infty}&=& \O\left(
\frac{\gamma}{(1-\gamma)^3}\left(c+\frac{1}{\lambda}\sqrt{d\log(\frac{Jc}{(1-\gamma)\delta})}\right)\sqrt{J^{\frac{1}{\beta_p}-1}\log^{1-\frac{1}{\beta_p}}(J) }
\right) +\O\left(\frac{\gamma^{L}}{(1-\gamma)^2} \right).
\end{eqnarray}

We choose $J$ and $L$ large enough so that each term on the right hand side is bounded by $\epsilon/2$. The choices of
\begin{eqnarray}\nonumber
J&=&\O\left(
\left(\frac{1}{\epsilon}\right)^{\frac{2\beta_p}{\beta_p-1}}\frac{\gamma^{\frac{2\beta_p}{\beta_p-1}}}{(1-\gamma)^{\frac{6\beta_p}{\beta_p-1}}}
\left(c+\frac{1}{\lambda}\sqrt{d\log(\frac{c}{(1-\gamma)\delta})}\right)^\frac{2\beta_p}{\beta_p-1}
\log^{\frac{\beta_p}{\beta_p-1}}\left(\frac{1}{\epsilon(1-\gamma)}\right)
\right),\\
L&=&\O\left(\frac{\log(\frac{1}{\epsilon})+\log(\frac{1}{1-\gamma})}{(1-\gamma)}\right),
\end{eqnarray}

with proper constants ensures $\|V^{\pi}-V^\ast\|_{\infty}\le \epsilon$. The expression can be simplified as $J=\tilde{\O}\left(\frac{\left(\log(\frac{1}{\delta})\right)^{\frac{\beta_p}{\beta_p-1}}}{\epsilon^{\frac{2\beta_p}{\beta_p-1}}\left(1-\gamma\right)^{\frac{6\beta_p}{\beta_p-1}}}\right)$ and $L=\tilde{\O}\left(\frac{1}{1-\gamma}\right)$, omitting the logarithmic and constant terms. That leads to 
\begin{eqnarray}
N=\tilde{\O}\left(\frac{\left(\log(\frac{1}{\delta})\right)^{\frac{\beta_p}{\beta_p-1}}}{\epsilon^{\frac{2\beta_p}{\beta_p-1}}\left(1-\gamma\right)^{\frac{7\beta_p-1}{\beta_p-1}}}\right).
\end{eqnarray}

\paragraph{Exponential Eigendecay:} Consider a kernel with $(C_{e_1}, C_{e_2}, \beta_e)$ polynomial eigendecay. We have the following bound on $\Gamma_{K,\lambda}$. For all $J\in\mathbb{N}$,
\begin{eqnarray}
\Gamma_{K,\lambda}(J) = \O(\log^{1+\frac{1}{\beta_e}}(J))
\end{eqnarray}

We replace this bound in the error in the value function of $\pi$ obtained by KQLearn, in Theorem~\ref{thm:main}, and obtain
\begin{eqnarray}\nonumber
\left\|V^{\pi}-V^\ast\right\|_{\infty}&=& \O\left(
\frac{\gamma}{(1-\gamma)^3}\left(c+\frac{1}{\lambda}\sqrt{d\log(\frac{Jc}{(1-\gamma)\delta})}\right)\sqrt{\frac{\log^{1+\frac{1}{\beta_e}}(J)}{J} }
\right) +\O\left(\frac{\gamma^{L-1}}{(1-\gamma)^2} \right).
\end{eqnarray}

We choose $J$ and $L$ large enough so that each term on the right hand side is bounded by $\epsilon/2$. The choices of
\begin{eqnarray}\nonumber
J&=&\O\left(
\left(\frac{1}{\epsilon}\right)^{2}\frac{\gamma^{2}}{(1-\gamma)^{6}}
\left(c+\frac{1}{\lambda}\sqrt{d\log(\frac{c}{(1-\gamma)\delta})}\right)^2
\log^{2+\frac{1}{\beta_e}}\left(\frac{1}{\epsilon(1-\gamma)}\right)
\right),\\
L&=&\O\left(\frac{\log(\frac{1}{\epsilon})+\log(\frac{1}{1-\gamma})}{(1-\gamma)}\right),
\end{eqnarray}

with proper constants ensures $\|V^{\pi}-V^\ast\|_{\infty}\le \epsilon$. The expression can be simplified as $J=\tilde{\O}\left(\frac{\log(\frac{1}{\delta})}{\epsilon^{2}\left(1-\gamma\right)^{6}}\right)$ and $L=\tilde{\O}\left(\frac{1}{1-\gamma}\right)$, omitting the logarithmic and constant terms. That leads to 
\begin{eqnarray}
N=\tilde{\O}\left(\frac{\log(\frac{1}{\delta})}{\epsilon^{2}\left(1-\gamma\right)^{7}}\right).
\end{eqnarray}

\section{Proof of Lemmas}
\label{appx:prooflem}

In this section, we provide the proof of auxiliary lemmas. 

\subsection{Proof of Lemma \ref{lem:RKHS_PV} [RKHS norm of $PV$]}

We have

\begin{eqnarray}\nonumber
\|PV\|_{\H_K} &=& \left\|\int_{s'\in\S}P(s'|\cdot,\cdot)V(s')ds' \right\|_{\H_K}\\\nonumber
&\le& \int_{s'\in\S}\left\|P(s'|\cdot,\cdot)V(s') \right\|_{\H_K}ds'\\\nonumber
&=&\int_{s'\in\S}\left\|P(s'|\cdot,\cdot)\right\|_{\H_K}V(s') ds'\\
&\le&\int_{s'\in\S}V(s') ds'
\end{eqnarray}

where the last inequality holds by Assumption~\ref{asp:RKHSnormP}. We note that $\int_{s'\in\S}V(s') ds' \le \frac{1}{1-\gamma}\int_{s'}ds'\le \frac{c}{1-\gamma}$ where $c$ is the volume of $\S$. 

\subsection{Proof of Lemma~\ref{lem:sub-Gauss_PV} [Sub-Gaussianity of $PV$]}
The first part, $\mathbb{E}\left[[\widehat{P}V](s,a)\right]=PV(s,a)$, follows from the definition of $\widehat{P}V$. For the second part note that $\widehat{P}V$ is a random variable with a bounded support in $[0,\frac{1}{1-\gamma}]$ by definition. Hoeffding lemma states that: let $X$ be any random variable such that $0\le X\le a$, then for any $\zeta\in\mathbb{R}$, $\mathbb{E}\left[\exp\left(\zeta(X-\mathbb{E}[X])\right)\right]\le \exp(\frac{\zeta^2 a^2}{8})$. Applying Hoeffding lemma, we can see that $\widehat{P}V$ is sub-Gaussian with parameter $\frac{1}{2(1-\gamma)}$.

\subsection{Proof of Lemma~\ref{lem:valerr} [Error in value function based on the error in proxy Q-function]}

The proof follows similar steps as the proof of Lemma 1.11 in \cite{agarwal2019} which considered finite state-actions. First, recall the following definitions from Section \ref{sec:prelim},
\begin{equation*}
Q^{\pi_{\tilde{Q}}}(s, a)=r(s, a)+\gamma  [PV^{\pi_{\tilde{Q}}}](s,a)\quad\text{and}\quad Q^\ast(s,a)=\max_{\pi}Q(s,\pi(s))
\end{equation*}

Fix state $s\in\S$ and let $\tilde{a}_s=\pi_{\tilde{Q}}(s)$. We have
\begin{equation*}
\begin{aligned}
V^{\ast}(s)-V^{\pi_{\tilde{Q}}}(s)=& Q^{\ast}\left(s, \pi^{\ast}(s)\right)-Q^{\pi_{\tilde{Q}}}(s, \tilde{a}_s) \\
=& Q^{\ast}\left(s, \pi^{\ast}(s)\right)-Q^{\ast}(s, \tilde{a}_s)+Q^{\ast}(s, \tilde{a}_s)-Q^{\pi_{\tilde{Q}}}(s, \tilde{a}_s) \\
=& Q^{\ast}\left(s, \pi^{\ast}(s)\right)-Q^{\ast}(s, \tilde{a}_s)+\gamma[PV^{\ast}](s, \tilde{a}_s)-\gamma[PV^{\pi_{\tilde{Q}}}](s, \tilde{a}_s) \\
=& Q^{\ast}\left(s, \pi^{\ast}(s)\right)-Q^{\ast}(s, \tilde{a}_s)+\gamma \mathbb{E}_{s^{\prime} \sim P(\cdot \mid s, \tilde{a}_s)}\left[V^{\ast}\left(s^{\prime}\right)-V^{\pi_{\tilde{Q}}}\left(s^{\prime}\right)\right] \\
\leq & Q^{\ast}\left(s, \pi^{\ast}(s)\right)-\tilde{Q}\left(s, \pi^{\ast}(s)\right)+\tilde{Q}(s, \tilde{a}_s)-Q^{\ast}(s, \tilde{a}_s) \\
&+\gamma \mathbb{E}_{s^{\prime} \sim P(s, \tilde{a}_s)}\left[V^{\ast}\left(s^{\prime}\right)-V^{\pi_{\tilde{Q}}}\left(s^{\prime}\right)\right] \\
\leq & 2\left\|\tilde{Q}-Q^{\ast}\right\|_{\infty}+\gamma\left\|V^{\ast}-V^{\pi_{\tilde{Q}}}\right\|_{\infty},
\end{aligned}
\end{equation*}
where the first inequality uses $\tilde{Q}\left(s, \pi^{\ast}(s)\right) \leq \tilde{Q}(s, \tilde{a}_s)$ by definition of $\tilde{a}_s$. 
We thus have
\begin{eqnarray}
\left\|V^{\ast}-V^{\pi_{\tilde{Q}}}\right\|_{\infty} \le 2\left\|\tilde{Q}-Q^{\ast}\right\|_{\infty}+\gamma\left\|V^{\ast}-V^{\pi_{\tilde{Q}}}\right\|_{\infty}. 
\end{eqnarray}

Rearranging this inequality, we arrive at the lemma
\begin{eqnarray}
\left\|V^{\ast}-V^{\pi_{\tilde{Q}}}\right\|_{\infty} \le \frac{2}{1-\gamma}\left\|\tilde{Q}-Q^{\ast}\right\|_{\infty}. 
\end{eqnarray}

\end{document}